\documentclass[sigconf]{acmart}
\usepackage{wrapfig}
\usepackage{multirow}
\setcopyright{none}
\renewcommand\footnotetextcopyrightpermission[1]{}
\AtBeginDocument{\fancypagestyle{standardpagestyle}{\fancyhf{}}\fancypagestyle{firstpagestyle}{\fancyhf{}}\pagestyle{standardpagestyle}\thispagestyle{firstpagestyle}}
\AtBeginDocument{%
  }

\begin{document}

\title{IsaacIPC: Coupling High-Fidelity Simulation and Realistic Rendering for Contact-Rich Robotic Systems}



\author{Qixin Liang}
\affiliation{%
  \institution{Anker Humanoid Lab}
  \city{Shenzhen}
  \country{China}
}
\affiliation{%
  \institution{The University of Hong Kong}
  \city{Hong Kong SAR}
  \country{China}
}

\author{Zhongqing Han}
\affiliation{%
  \institution{Anker Humanoid Lab}
  \city{Shenzhen}
  \country{China}
}


\begin{abstract}
We present IsaacIPC, a robotic simulation framework that couples GPU accelerated incremental
potential contact (IPC) with Isaac Sim\,/\,Isaac Lab.
IsaacIPC maps simulated deformation between simulation and visual meshes,
enabling real-time realistic rendering with applications to data collection and
policy evaluation.
For tactile sensing, we introduce the geometric mortar contact potential
(GMCP), which defines a barrier potential over contact samples on tactile
surfaces to better resolve contact-pressure distributions.
We evaluate GMCP on contact benchmarks and demonstrate IsaacIPC on
rigid--deformable robotic simulations including a quadruped robot, a dexterous
hand, and a universal manipulation interface (UMI) gripper.
\end{abstract}


\begin{teaserfigure}
  \includegraphics[width=\textwidth]{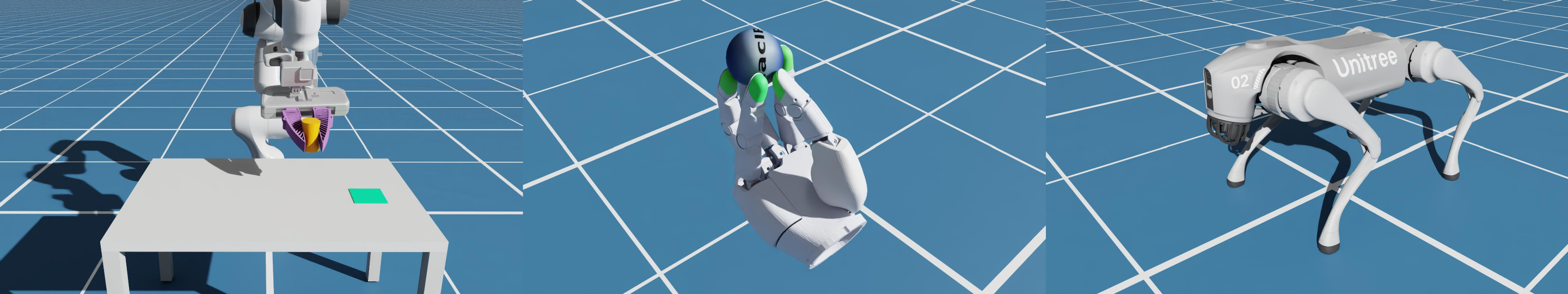}
  \caption{IsaacIPC qualitative demonstrations across rigid--deformable robotic systems: a soft Universal Manipulation Interface (UMI) gripper, a dexterous hand, and a quadruped robot.}
  \Description{Three qualitative IsaacIPC demonstrations showing a universal manipulation interface (UMI) gripper, a dexterous hand, and a quadruped robot.}
  \label{fig:teaser}
\end{teaserfigure}

\maketitle

\section{Introduction}

High-fidelity robotic simulation frameworks are increasingly important for
embodied intelligence: they enable scalable policy training, systematic
evaluation, and contact-rich interactions that are expensive to collect in the
real world~\cite{Genesis,mittal2025isaac}.
Contact simulation is central to these systems because manipulation, locomotion, and tactile sensing all rely heavily on physical interaction with the environment.
Incremental Potential Contact (IPC)~\cite{Li2020IPC} is attractive in this
setting because its barrier formulation, combined with continuous collision
detection, robustly maintains intersection- and inversion-free simulation under large
deformations and challenging contact.

Recent work has improved IPC along complementary directions.
For efficiency, IPC has been accelerated by projective dynamics~\cite{Lan2022PDGPU},
preconditioned nonlinear conjugate gradients~\cite{Shen2024PNCGIPC}, and
GPU Gauss--Newton solvers~\cite{Huang2024GIPC,Huang2025StiffGIPC}.
For accuracy, Convergent IPC~\cite{Li2023ConvergentIPC} and GCP~\cite{Huang2025GCP}
study continuous contact potentials and their discretization.
These developments have made IPC-based contact simulation increasingly useful in robotics:
IPC-GraspSim~\cite{Kim2022IPCGraspSim} applies IPC to grasp simulation, while
TacIPC~\cite{Du2024TacIPC}, Taccel~\cite{Li2025Taccel}, TacEx~\cite{Nguyen2024TacEx},
UniVTAC~\cite{Chen2026UniVTAC}, and Tac2Real~\cite{Yan2026Tac2Real} use IPC or
its variants for tactile simulation, data generation, and policy learning.

Tactile sensing places a strong requirement on contact pressure accuracy, as many tactile sensors rely on contact-induced deformation to generate signals.
TacEx~\cite{Nguyen2024TacEx} integrates GIPC~\cite{Huang2024GIPC} with
Isaac Sim\,/\,Isaac Lab~\cite{mittal2025isaac} for GelSight
visuotactile sensor~\cite{yuan2015measurement}, but
GIPC-modeled objects are not mapped to high-fidelity visual assets with
textures and material appearance for NVIDIA Omniverse~\cite{NVIDIAOmniverse2026}.
UniVTAC~\cite{Chen2026UniVTAC} builds on TacEx for visuo-tactile data generation
and benchmarking, and inherits this limitation.
This leaves a practical gap for real-to-sim workflows that require robust
deformable contact, scalable rollout, and high-fidelity rendering from detailed
visual assets.

In this work, we make two key contributions.
\begin{itemize}
    \item First, we introduce IsaacIPC, a simulation framework that
    bridges IsaacSim/Lab~\cite{mittal2025isaac} with libuipc~\cite{libuipc}
    through a dual-mesh mapper,
    combining high-fidelity contact simulation with realistic rendering for
    data collection and robot-learning policy evaluation.
    \item Second, we propose the Geometric Mortar Contact Potential (GMCP),
    a mortar-based contact formulation that samples contact on tactile surfaces
    and incorporates the barrier potential to improve contact-pressure
    simulation for tactile sensing.
\end{itemize}
Together, these contributions offer a practical foundation for improving contact-rich simulation in robotic interactions.

\section{Related Work}

\subsection{Barrier-Based Contact Simulation}

Incremental Potential Contact (IPC)~\cite{Li2020IPC} formulates contact as a smoothed barrier potential with continuous collision detection (CCD), guaranteeing intersection- and inversion-free trajectories under implicit time integration.
This framework has been extended along multiple axes: to rigid bodies with curved-trajectory CCD~\cite{Ferguson2021RigidIPC}, to stiff affine bodies with reduced degrees of freedom~\cite{Lan2022ABD}, to unified rigid-deformable multibody systems with articulation constraints~\cite{Chen2022UnifiedIPC}, to high-order finite element meshes~\cite{Ferguson2023HighOrderIPC}, and to adaptive remeshing within the timestep~\cite{Ferguson2023ITR}.

A major thrust of recent work accelerates IPC to practical speeds.
Lan et al.~\cite{Lan2022PDGPU} integrate the barrier into projective dynamics on the GPU.
Shen et al.~\cite{Shen2024PNCGIPC} eliminate Hessian assembly via a preconditioned nonlinear conjugate gradient (PNCG) method, and Zhang et al.~\cite{Zhang2026MPNCGIPC} further improve this with multilevel preconditioning.
GIPC~\cite{Huang2024GIPC} derives analytic eigensystems for GPU-friendly Gauss--Newton solves, and StiffGIPC~\cite{Huang2025StiffGIPC} adds multilevel preconditioning and affine-deformable coupling for stiff materials.
Barrier-Augmented Lagrangian~\cite{Guo2024BarrierAL} combines barriers with augmented Lagrangian for improved conditioning, while AL-IPC~\cite{Zheng2026ALIPC} replaces barriers entirely with a second-order augmented Lagrangian, avoiding TOI locking.

On the accuracy side, Convergent IPC~\cite{Li2023ConvergentIPC} reformulates IPC in the continuous setting and gives a convergent discretization, with convergence demonstrated under joint refinement of the mesh, time step, and contact activation distance.
The Geometric Contact Potential (GCP)~\cite{Huang2025GCP} derives a continuum barrier from geometric interaction sets and uses an adaptive locality parameter to avoid rest-state spurious forces while relaxing the coupling between barrier extent and mesh resolution.
Classical mortar contact methods improve interface consistency by enforcing contact constraints in a weak, integrated form over overlapping slave-master regions, and are widely used for accurate contact pressure transfer on nonmatching meshes~\cite{Wohlmuth2000DualMortar,Puso2004Mortar,farah2018mortar}.
They are typically formulated with Lagrange multiplier or augmented Lagrangian enforcement rather than feasible-iterate barrier formulations~\cite{Huang2025GCP}; our work instead incorporates mortar integration into a barrier potential to improve contact pressure accuracy while retaining non-penetration.

\subsection{IPC in Robotic Simulation}

IPC has recently been adopted in robotics applications involving robust contact handling and deformable objects.
IPC-GraspSim~\cite{Kim2022IPCGraspSim} introduced IPC to parallel-jaw grasp simulation with compliant fingertips, showing that explicitly modeling soft jaw deformation and frictional contact can improve grasp outcome prediction.
ZeMa~\cite{Du2024ZeMa} further extended IPC-based simulation to unified rigid-deformable robotic manipulation, combining (finite element method) FEM and affine rigid bodies for applications including grasp generation, grasp repair, and reinforcement learning.
Embedded IPC~\cite{Du2024EmbeddedIPC} improves the runtime of this line of work by using a reduced subspace for elasticity while retaining contact constraints on an embedded high-resolution surface, targeting interactive robot manipulation scenarios.

IPC-based methods have also been explored for visuo-tactile sensing, where elastomer deformation and frictional contact directly determine the tactile signal.
TacIPC~\cite{Du2024TacIPC} uses FEM with IPC to simulate optical tactile sensor gels with intersection- and inversion-free deformation, enabling tactile image and marker displacement prediction.
TacEx~\cite{Nguyen2024TacEx} integrates GIPC~\cite{Huang2024GIPC} with Isaac Sim\,/\,Isaac Lab~\cite{mittal2025isaac}, combining FEM with visuotactile rendering modules for GelSight sensors~\cite{yuan2015measurement}.
Taccel~\cite{Li2025Taccel} scales this direction further by combining IPC and affine body dynamics with NVIDIA Warp~\cite{NVIDIAWarp} to support many parallel tactile-robotics environments.
UniVTAC~\cite{Chen2026UniVTAC} builds on TacEx~\cite{Nguyen2024TacEx} from the data and evaluation side, providing a unified visuo-tactile simulation platform, representation-learning pipeline, and benchmark for tactile-driven manipulation.
Tac2Real~\cite{Yan2026Tac2Real} targets online reinforcement learning and zero-shot sim-to-real transfer, using PNCG-IPC~\cite{Shen2024PNCGIPC} as the contact solver in a visuotactile simulation framework.
These works demonstrate the value of IPC-quality contact for robotics, while also motivating simulation frameworks that combine robust deformable contact, scalable GPU execution, and direct integration with modern robot-learning environments.

\section{IsaacIPC}\label{sec:isaacipc}

In this section, we present IsaacIPC (Fig.~\ref{fig:isaacipc_arch}), a simulation framework that couples the high-fidelity real-time rendering and massively parallel environment infrastructure of Isaac Sim\,/\,Isaac Lab~\cite{mittal2025isaac} with the GPU-accelerated contact simulation of libuipc~\cite{libuipc,Huang2025StiffGIPC,Huang2024GIPC}, incorporating our GMCP contact model (\S\ref{sec:mortar}) and a Dual-Mesh Mapper (\S\ref{sec:dual_mesh}).

\subsection{System Architecture}\label{sec:arch}
\begin{figure}[t]
  \centering
  \includegraphics[width=\linewidth]{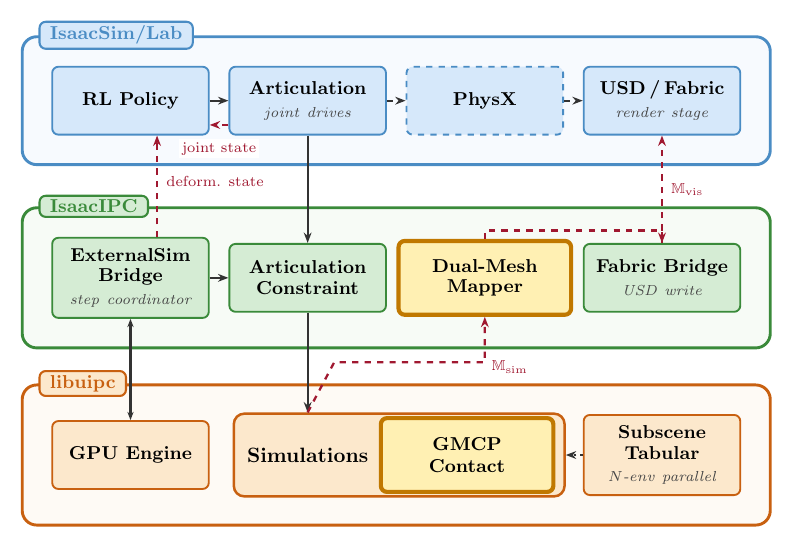}
  \caption{System architecture of IsaacIPC.}
  \label{fig:isaacipc_arch}
\end{figure}

IsaacIPC adopts a three-layer architecture (Fig.~\ref{fig:isaacipc_arch}).
The \emph{IsaacSim/Lab} layer provides robot articulation and photorealistic rendering via the Universal Scene Description (USD)\,/\,Fabric stage, with an optional one-way coupling to PhysX~\cite{NVIDIA_PhysX}.
The \emph{IsaacIPC} layer mediates between IsaacSim/Lab and libuipc: the \emph{ExternalSim Bridge} coordinates each physics step; the \emph{Articulation Constraint} reads joint-driven body poses and applies them as constraints inside libuipc; the \emph{Dual-Mesh Mapper} (\S\ref{sec:dual_mesh}) interpolates the visual mesh $\mathbb{M}_{\mathrm{vis}}$ onto the simulation mesh $\mathbb{M}_{\mathrm{sim}}$ each frame; and the \emph{Fabric Bridge} writes the interpolated $\mathbb{M}_{\mathrm{vis}}$ vertex positions back to the USD\,/\,Fabric stage.
The \emph{libuipc} layer runs all physics on the GPU: the \emph{Simulations} block handles rigid and deformable body dynamics with GPU contact, of which GMCP (\S\ref{sec:mortar}) is one contact model; the \emph{Subscene Tabular} layout partitions the simulation into multiple fully isolated subscenes on a single GPU, enabling massively parallel rollouts.

\subsection{Dual-Mesh Mapper}\label{sec:dual_mesh}

\begin{wrapfigure}{r}{0.36\columnwidth}
  \vspace{-6pt}
  \centering
  \includegraphics[width=0.36\columnwidth]{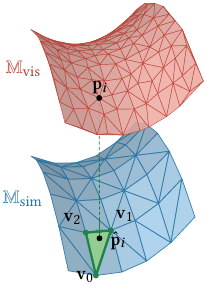}
  \vspace{-10pt}
  \caption{Dual-mesh mapper.}
  \label{fig:dual_mesh}
  \vspace{-6pt}
\end{wrapfigure}

Reconstructed or procedurally generated simulation assets~\cite{lai2025hunyuan3d25} typically provide only surface meshes, whereas deformable simulation requires volumetric discretization (e.g., tetrahedralization via TetGen~\cite{Si2015TetGen}), which may alter mesh topology and break texture-coordinate parameterization and material bindings essential for photorealistic rendering.
Inspired by barycentric embedding in skeleton-driven animation~\cite{Capell2002Skeleton} and Embedded IPC~\cite{Du2024EmbeddedIPC}, we therefore maintain two meshes: a \emph{simulation mesh} $\mathbb{M}_\mathrm{sim}$ for contact and deformation, and a \emph{visual mesh} $\mathbb{M}_\mathrm{vis}$ for rendering, preserving the original texture coordinates, materials bindings, and topology.

\paragraph{Precomputation.}
At initialization, each visual vertex $\mathbf{p}_i \in \mathcal{V}$ ($i = 1,\dots,N$) is embedded onto the surface of the simulation mesh $\mathbb{M}_\mathrm{sim}$ via closest-point projection.
For each $\mathbf{p}_i$, we find the nearest simulation triangle $T_i \in \mathbb{M}_\mathrm{sim}$ (subscript $i$ labels the assignment to vertex $\mathbf{p}_i$, not a global enumeration of triangles) with vertices $\{\mathbf{v}_0,\mathbf{v}_1,\mathbf{v}_2\}$, and compute its barycentric coordinates $(u_i, v_i, w_i)$ such that
\begin{equation}
  \hat{\mathbf{p}}_i = u_i\,\mathbf{v}_0 + v_i\,\mathbf{v}_1 + w_i\,\mathbf{v}_2, \qquad u_i + v_i + w_i = 1,
  \label{eq:bary_proj}
\end{equation}
where $\hat{\mathbf{p}}_i$ is the projection of $\mathbf{p}_i$ onto $T_i$.
Let $\mathbf{e}_1 = \mathbf{v}_1-\mathbf{v}_0$, $\mathbf{e}_2 = \mathbf{v}_2-\mathbf{v}_0$, and $\mathbf{n}_i = (\mathbf{e}_1\times\mathbf{e}_2)/\|\mathbf{e}_1\times\mathbf{e}_2\|$ be the unit normal of $T_i$.
Setting $\mathbf{d} = \mathbf{p}_i - \mathbf{v}_0$ and $\delta_i = \mathbf{d}\cdot\mathbf{n}_i$ (signed offset of $\mathbf{p}_i$ from the plane of $T_i$), we have
\begin{equation}
  \mathbf{d} = v_i\,\mathbf{e}_1 + w_i\,\mathbf{e}_2 + \delta_i\,\mathbf{n}_i.
  \label{eq:decomp}
\end{equation}
Taking the dot product of Eq.~\eqref{eq:decomp} with $\mathbf{e}_1$ and $\mathbf{e}_2$ in turn, the $\delta_i\mathbf{n}_i$ term vanishes since $\mathbf{n}_i\cdot\mathbf{e}_1 = \mathbf{n}_i\cdot\mathbf{e}_2 = 0$, yielding
\begin{equation}
  \begin{pmatrix} \mathbf{e}_1\cdot\mathbf{e}_1 & \mathbf{e}_1\cdot\mathbf{e}_2 \\ \mathbf{e}_1\cdot\mathbf{e}_2 & \mathbf{e}_2\cdot\mathbf{e}_2 \end{pmatrix}
  \begin{pmatrix} v_i \\ w_i \end{pmatrix}
  =
  \begin{pmatrix} \mathbf{d}\cdot\mathbf{e}_1 \\ \mathbf{d}\cdot\mathbf{e}_2 \end{pmatrix},
  \label{eq:gram}
\end{equation}
with $u_i = 1 - v_i - w_i$.
The mapping $\{(T_i,\,u_i,\,v_i,\,w_i,\,\delta_i)\}_{i=1}^{N}$ is computed once and stored.

\paragraph{Per-frame update.}
After each simulation step, the simulation vertices are updated to deformed positions.
The visual vertex positions are recovered by interpolating the deformed simulation vertices and restoring the normal offset:
\begin{equation}
  \mathbf{p}_i' = u_i\,\mathbf{v}_0' + v_i\,\mathbf{v}_1' + w_i\,\mathbf{v}_2' + \delta_i\,\mathbf{n}_i', \qquad i = 1,\dots,N,
  \label{eq:bary_interp}
\end{equation}
where $\mathbf{v}_0', \mathbf{v}_1', \mathbf{v}_2'$ are the deformed positions of the three vertices of $T_i$, and $\mathbf{n}_i'$ is the unit normal of the deformed triangle $T_i$.
The mesh topology, texture coordinates, and material bindings of $\mathbb{M}_\mathrm{vis}$ remain unchanged throughout; only the vertex positions $\{\mathbf{p}_i'\}$ are updated each frame.

The mapping is rigid-invariant and introduces a piecewise-linear approximation error of $O(h^2)$ for general deformations, where $h$ is the simulation triangle edge length; rendering fidelity thus improves quadratically with mesh refinement.

\section{Geometric Mortar Contact Potential}\label{sec:method}

We consider contacts in which pressure is evaluated on a tactile surface.
We use this surface as the slave geometry.
The opposing geometry is the master side, whose local contact feature may be a
surface patch, a line segment, or a point.

\subsection{Contact Sampling}\label{sec:mortar}

\begin{figure}[t]
  \centering
  \includegraphics[width=\linewidth]{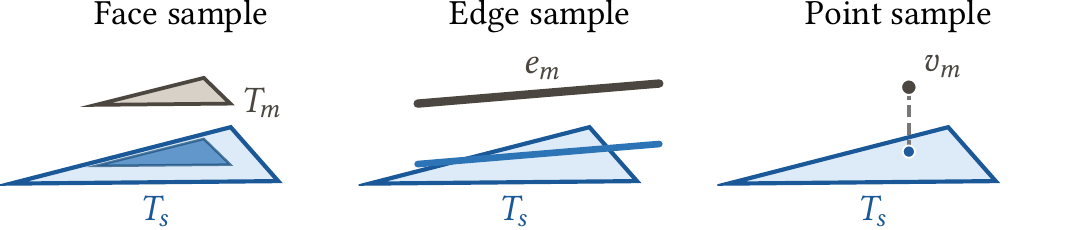}
  \caption{Contact sampling on a slave triangle, which is extracted from the tactile surface.}
  \label{fig:mortar_integration}
\end{figure}

The slave triangulated mesh $\mathcal{S}$ provides the local parameterization and normal
field $\mathbf{n}_s$.
Contact samples are constructed by projecting and clipping master features
against local slave triangles.
We refer to samples on clipped face overlaps as \emph{face samples}, samples on
clipped edge segments as \emph{edge samples}, and projected master vertices as
\emph{point samples} (see Fig.~\ref{fig:mortar_integration}).
Each sample $k$ is either a quadrature point on a clipped face overlap or
edge segment, or a projected master vertex.
It stores a position $\mathbf{x}_{s,k}$ on $\mathcal{S}$, an associated master
position $\mathbf{x}_{m,k}$, a sample weight $w_k$, and a signed gap $g_k$
measured along the slave normal.

\emph{Face samples.}
For a slave triangle $T_s$ and a master triangle $T_m$, we project both
triangles onto the tangent plane of $T_s$ and clip their overlap using
Sutherland-Hodgman clipping~\cite{Sutherland1974}.
The clipped polygon is fan-triangulated and integrated with 2D Gauss quadrature.
At sample $k$, barycentric coordinates $\beta_{a,k}^s$ and $\beta_{a,k}^m$
interpolate the slave and master positions,
\begin{equation}
  \mathbf{x}_{s,k} = \textstyle\sum_a \beta_{a,k}^s\mathbf{x}_{s,a},
  \qquad
  \mathbf{x}_{m,k} = \textstyle\sum_a \beta_{a,k}^m\mathbf{x}_{m,a}
  \qquad a=0,1,2.
\end{equation}
The signed gap is
\begin{equation}
  g_k = \mathbf{n}_s \cdot
  (\mathbf{x}_{m,k}-\mathbf{x}_{s,k}).
  \label{eq:gap}
\end{equation}

\emph{Edge samples.}
For a master edge $e_m=(\mathbf{x}_{m,0}^e,\mathbf{x}_{m,1}^e)$, we project the
edge onto the tangent plane of a slave triangle and clip the projected segment
by the triangle.
Using the edge parameter $\eta\in[0,1]$,
\begin{equation}
  \mathbf{x}_{m,k}=(1-\eta_k)\mathbf{x}_{m,0}^e
  +\eta_k\mathbf{x}_{m,1}^e,
\end{equation}
each 1D Gauss point gives slave barycentric coordinates
$\beta_{a,k}^s$ on $T_s$ and
\begin{equation}
  \mathbf{x}_{s,k}=\textstyle\sum_a \beta_{a,k}^s\mathbf{x}_{s,a},
  \qquad
  a = 0, 1, 2.
  \label{eq:edge_pos}
\end{equation}
The signed gap is
\begin{equation}
  g_k = \mathbf{n}_s \cdot
  (\mathbf{x}_{m,k}-\mathbf{x}_{s,k}).
  \label{eq:gap_edge}
\end{equation}

\emph{Point samples.}
For a master vertex, we project this vertex onto a slave triangle and generate
a point sample.
The sample uses the projected slave position $\mathbf{x}_{s,k}$ and the master vertex position $\mathbf{x}_{m,k}$ to compute the gap $g_k$ as in
Eq.~\eqref{eq:gap} and Eq.~\eqref{eq:gap_edge}.

\subsection{GMCP Barrier Potential}\label{sec:barrier}

Let $\mathcal{K}_q$ denote the contact samples of type $q$ on the slave surface.
The discrete barrier potential is
\begin{equation}
  \Psi = \sum_{q\in\{2,1,0\}}\sum_{k\in\mathcal{K}_q}
  \kappa_q\,w_k\,\gamma_k\,B(g_k,\varepsilon_k),
  \label{eq:gmcp_potential}
\end{equation}
where $q=2,1,0$ denotes face, edge, and point samples, $w_k$ is the sample
weight, with units corresponding to area, length, or a point contribution,
$\kappa_q$ is the barrier stiffness for sample type $q$, $\gamma_k$ is a
master-feature weight defined in Sec.~\ref{sec:direction}, and
$\varepsilon_k$ is the barrier support radius defined in
Sec.~\ref{sec:adaptive_eps}.
The barrier function $B$ follows IPC~\cite{Li2020IPC}; other function, such as GCP~\cite{Huang2025GCP}, can also be adopted.

For the current contact samples, we use the gradient and Hessian approximation
\begin{align}
  \nabla\Psi &= \sum_{q}\sum_{k\in\mathcal{K}_q} \kappa_q w_k\gamma_k
  \frac{\partial B}{\partial g_k}\nabla g_k,\\
  \nabla^2\Psi &= \sum_{q}\sum_{k\in\mathcal{K}_q} \kappa_q w_k\gamma_k
  \frac{\partial^2 B}{\partial g_k^2}
  \nabla g_k(\nabla g_k)^T,
\end{align}
with Hessian contributions projected to the positive semi-definite cone.

\subsection{Master-Feature Weights \texorpdfstring{$\gamma$}{gamma}}\label{sec:direction}

The weight $\gamma$ controls transitions between face, edge, and point samples
on the master side.
We use the $C^1$ Hermite step to smoothly scale sample contributions near
master-feature boundaries.
\begin{equation}
  H(x,\delta)=
  \begin{cases}
    0 & x\le 0,\\
    3t^2-2t^3,\quad t=x/\delta & 0<x<\delta,\\
    1 & x\ge \delta,
  \end{cases}
  \label{eq:smoothstep}
\end{equation}
For face samples, we use
\begin{equation}
  f_T = H\!\left(\min_a \beta_a^m,\delta_T\right),
\end{equation}
where $\beta_a^m$ are barycentric coordinates on the master triangle and
$\delta_T$ is the transition width near master-triangle edges.
For edge samples, we use
\begin{equation}
  f_e = H(\eta,\delta_e)H(1-\eta,\delta_e).
\end{equation}
Here $\eta$ is the master edge parameter and $\delta_e$ is the transition width
near master-edge endpoints.
The master-feature weight is
\begin{equation}
  \gamma =
  \begin{cases}
    f_T & \text{face sample},\\
    f_e & \text{edge sample},\\
    1   & \text{point sample}.
  \end{cases}
  \label{eq:gamma}
\end{equation}
Thus face samples decay near master-triangle edges, edge samples decay near
master-edge endpoints, and point samples keep full weight.

\subsection{Adaptive Barrier Support Radius}\label{sec:adaptive_eps}

Inspired by GCP~\cite{Huang2025GCP}, we use an adaptive support radius $\varepsilon_k$ for each contact sample $k$ to avoid spurious forces.
We assign the support radius as:
\begin{equation}
  \varepsilon_k =
  \min\!\big(0.9\,g_{\mathrm{ref},k},\varepsilon_{\mathrm{max}}\big),
  \label{eq:adaptive_eps}
\end{equation}
where $g_{\mathrm{ref},k}$ is the gap to the associated master
feature, and $\varepsilon_{\mathrm{max}}$ is a prescribed support-radius bound.
This choice gives $g_{\mathrm{ref},k}>\varepsilon_k$ which avoids spurious forces.

\subsection{Linearized Gap Step Size}\label{sec:ccd}

For the current contact samples, we compute a step-size bound from the linearized
gap.
For a displacement $\Delta\mathbf{x}$ that decreases sample $k$'s gap, we use
\begin{equation}
  \alpha_k^* =
  0.9\,\frac{g_k}{-\nabla g_k\cdot\Delta\mathbf{x}},
  \qquad \nabla g_k\cdot\Delta\mathbf{x}<0.
  \label{eq:ccd}
\end{equation}
The global step size is
\begin{equation}
  \alpha = \min_k \alpha_k^*,
  \label{eq:global_ccd}
\end{equation}
where the minimum is taken over all contact samples $k$ with
$\nabla g_k\cdot\Delta\mathbf{x}<0$, with $g_k$ and $\nabla g_k$ evaluated at
the penetration-free positions.


\section{Results}\label{sec:results}

In this section, we evaluate GMCP and IsaacIPC through quantitative and qualitative tests. 
All tests were run on a workstation with an AMD EPYC 9T24 CPU with 32 cores and an NVIDIA RTX5880-Ada-48Q GPU.

\subsection{Quantitative Tests for GMCP}\label{sec:results_quant}

In this subsection, $E$ is Young's modulus, $\nu$ is Poisson's ratio, $u_x$, $u_y$, and $u_z$ are displacement components along the $x$, $y$, and $z$ axes, and $\sigma_{ij}$, $i,j \in \{x,y,z\}$ is a Cauchy stress component. 
All quantities are reported in SI units. 
GMCP is implemented in libuipc, which branches from \texttt{main} at commit \texttt{1dd8e36}.

\emph{Contact patch test.}
The contact patch test checks whether a contact method transmits a uniform stress state across a non-matching interface~\cite{wriggers2006computational}.
We use two linear-elastic blocks of size $1\times1\times0.5$, separated by an initial gap of $0.002$.
\begin{wrapfigure}{r}{0.4\columnwidth}
  \vspace{-6pt}
  \centering
  \includegraphics[width=0.4\columnwidth]{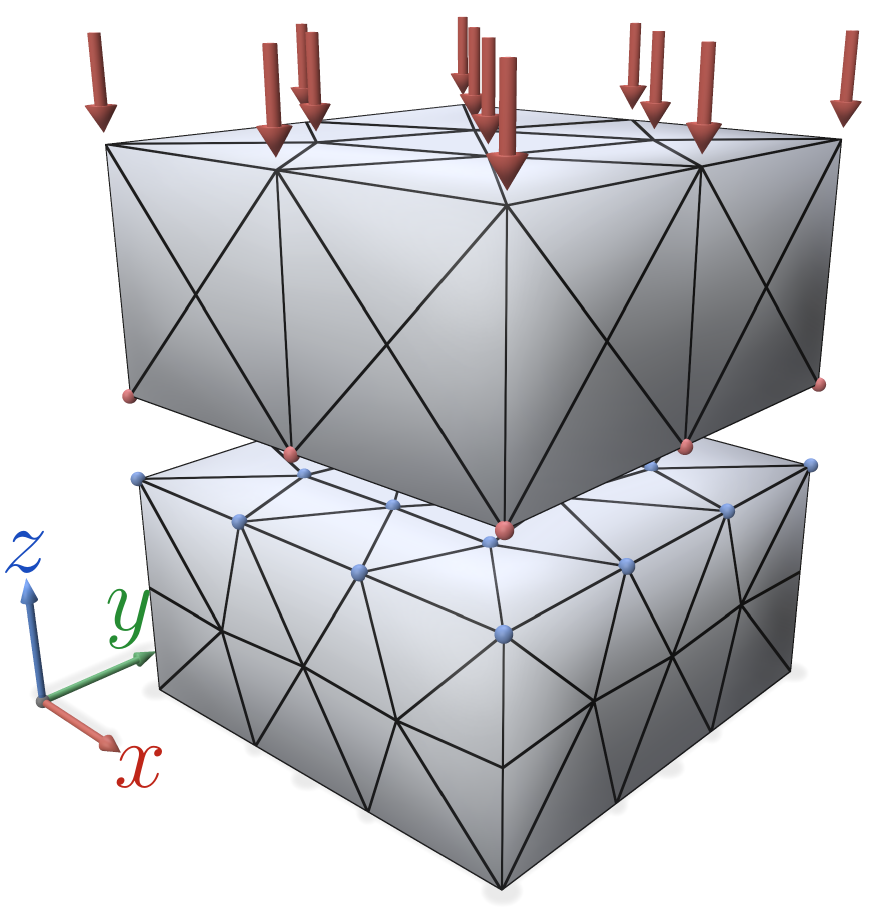}
  \vspace{-10pt}
  \caption{Contact patch test.}
  \label{fig:patch_test_setup}
  \vspace{-6pt}
\end{wrapfigure}
The bottom block has 56 vertices and 125 tetrahedra, while the top block has 32 vertices and 70 tetrahedra.
Both use $E=1000$ and $\nu=0$.
The bottom surface of the bottom block is constrained with $u_x=u_y=u_z=0$, the top block is constrained with $u_x = u_y = 0$, and a uniform pressure of $10$ is applied to the top surface of the top block along $-z$.
All runs are quasi-static, frictionless, and use barrier support radius $0.001$.

The analytical solution is $\sigma_{zz}=-10$ uniformly in both blocks, with all other stress components equal to zero.
We report the maximum relative $\sigma_{zz}$ error and the maximum spurious stress, denoted $\sigma_{\mathrm{spur}}$, defined as the largest absolute non-target Cauchy stress component among $\sigma_{xx}$, $\sigma_{yy}$, $\sigma_{xy}$, $\sigma_{yz}$, and $\sigma_{xz}$.
As shown in Table~\ref{tab:patch_test}, GMCP transmits the target uniform stress with small relative error for all tested barrier stiffness $\kappa$ values.

\begin{table}[h]
  \centering
  \caption{GMCP contact patch test results.}
  \label{tab:patch_test}
  \small
  \setlength{\tabcolsep}{6pt}
  \begin{tabular}{lcc}
    \toprule
    $\kappa$ & $\sigma_{zz}$ max rel.\ err. & $\sigma_{\mathrm{spur}}$ \\
    \midrule
    $10^4$ & $8.00\!\times\!10^{-4}$ & $6.00\!\times\!10^{-3}$ \\
    $10^6$ & $4.66\!\times\!10^{-5}$ & $2.48\!\times\!10^{-4}$ \\
    $10^8$ & $4.66\!\times\!10^{-5}$ & $2.48\!\times\!10^{-4}$ \\
    \bottomrule
  \end{tabular}
\end{table}

\emph{Hertzian Contact.}
The Hertzian contact test evaluates how accurately the method recovers the normal pressure distribution in smooth curved contact at finite resolution~\cite{laursen2003computational}.
We use a quarter-symmetry model in which a deformable one-eighth hemisphere of radius $R=0.05$ indents a deformable elastic block represented by a quarter-cylinder sector of outer radius $0.12$ and height $0.06$.
The block has 10,488 vertices and 55,921 tetrahedra, and the hemisphere has 11,838 vertices and 61,843 tetrahedra.
The minimum initial gap between the block and the hemisphere is $5\times10^{-5}$.
Both bodies use the same linear-elastic material, with $E=2.1\times10^{11}$ and $\nu=0.3$.
The block bottom is fixed with $u_z=0$.
\begin{wrapfigure}{r}{0.42\columnwidth}
  \centering
  \vspace{-6pt}
  \includegraphics[width=0.42\columnwidth]{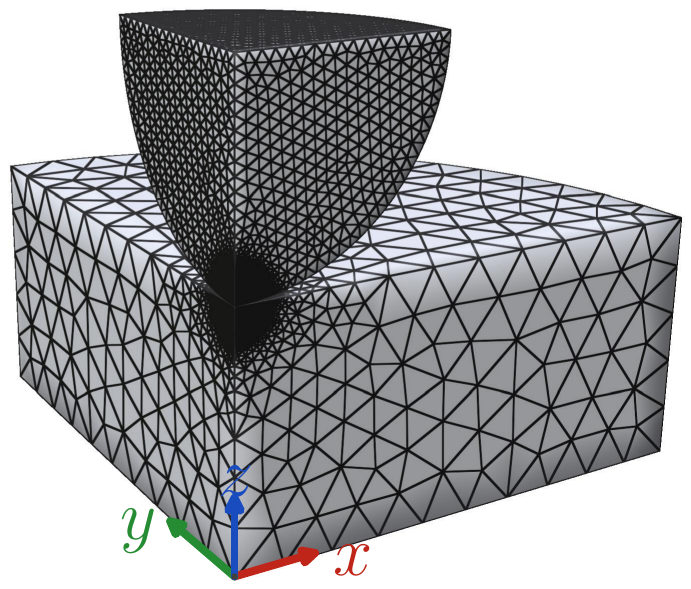}
  \caption{Hertzian contact.}
  \label{fig:hertz_setup}
  \vspace{-10pt}
\end{wrapfigure}
On both bodies, the two symmetry planes $x=0$ and $y=0$ impose $u_x=0$ and $u_y=0$, respectively.
On the hemisphere, a uniform pressure $Q=10^7$ is applied to the top surface along $-z$ and ramped over 10 load steps.
The simulation is quasi-static and frictionless, with GMCP barrier support radius $10^{-5}$.

\begin{figure}[]
  \centering
  \includegraphics[width=0.9\linewidth]{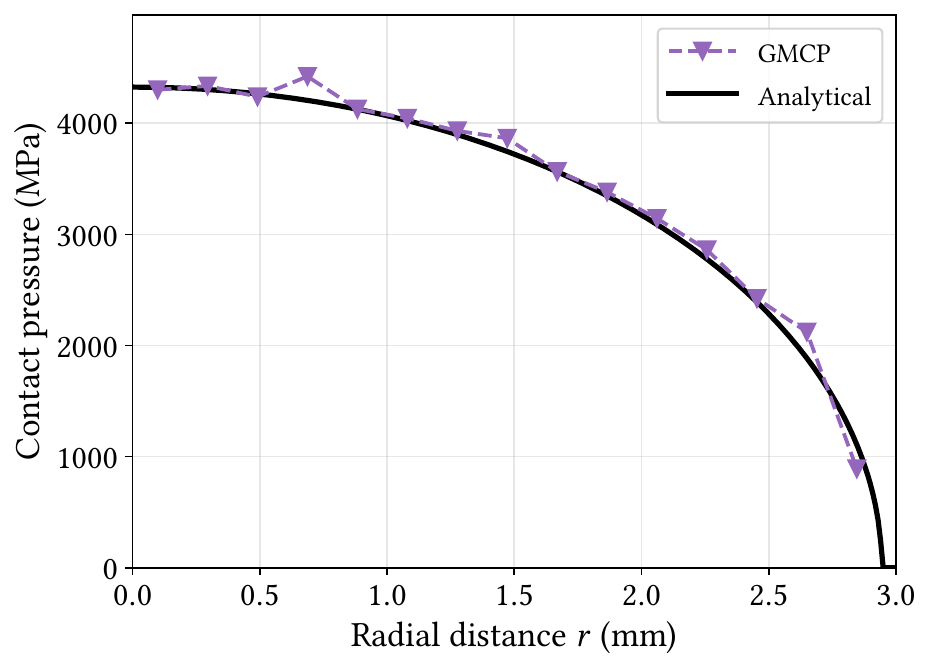}
  \caption{Hertz contact of GMCP compared with the analytical solution.}
  \label{fig:hertz_pressure}
\end{figure}

The analytical contact pressure is given by the Hertz solution~\cite{johnson1987contact},
\[
p(r)=\frac{3Q R^2}{2 \alpha^2}\sqrt{1-\frac{r^2}{\alpha^2}},
\qquad
\alpha =\left(\frac{3q\pi R^3}{4E^*}\right)^{1/3},
\]
where $r$ is the radial distance from the contact center and
$E^*=E/[2(1-\nu^2)]$ is the effective Young's modulus.
Fig.~\ref{fig:hertz_pressure} compares this analytical profile with the
numerical GMCP result, which shows close agreement. The small oscillations of
a few numerical points are expected to decrease with mesh refinement or
higher-order elements.

\subsection{Qualitative Tests for IsaacIPC}\label{sec:results_qual}
Unless otherwise specified in this subsection, rigid bodies are simulated using affine body dynamics~\cite{Lan2022ABD}, while deformable bodies are simulated using linear tetrahedral finite elements. This subsection focuses on qualitative demonstrations of the capabilities of the IsaacIPC framework. The contact model can be GMCP or another suitable model, and detailed simulation settings are omitted.

\emph{Legged locomotion.}
The Unitree Go2 quadruped robot~\cite{UnitreeRobotics2026}, shown in the right panel of Fig.~\ref{fig:teaser}, is modeled as a full-body articulated system with rigid links and joint constraints in libuipc. All components are represented as rigid bodies except for the four black foot pads, which are modeled using FEM. For simulation efficiency, the simulation mesh is coarser than the rendering mesh; nevertheless, high-quality visual rendering is preserved through the dual-mesh mapper (Fig.~\ref{fig:dual_mesh}).

\begin{figure}[t]
  \centering
  \includegraphics[width=\linewidth]{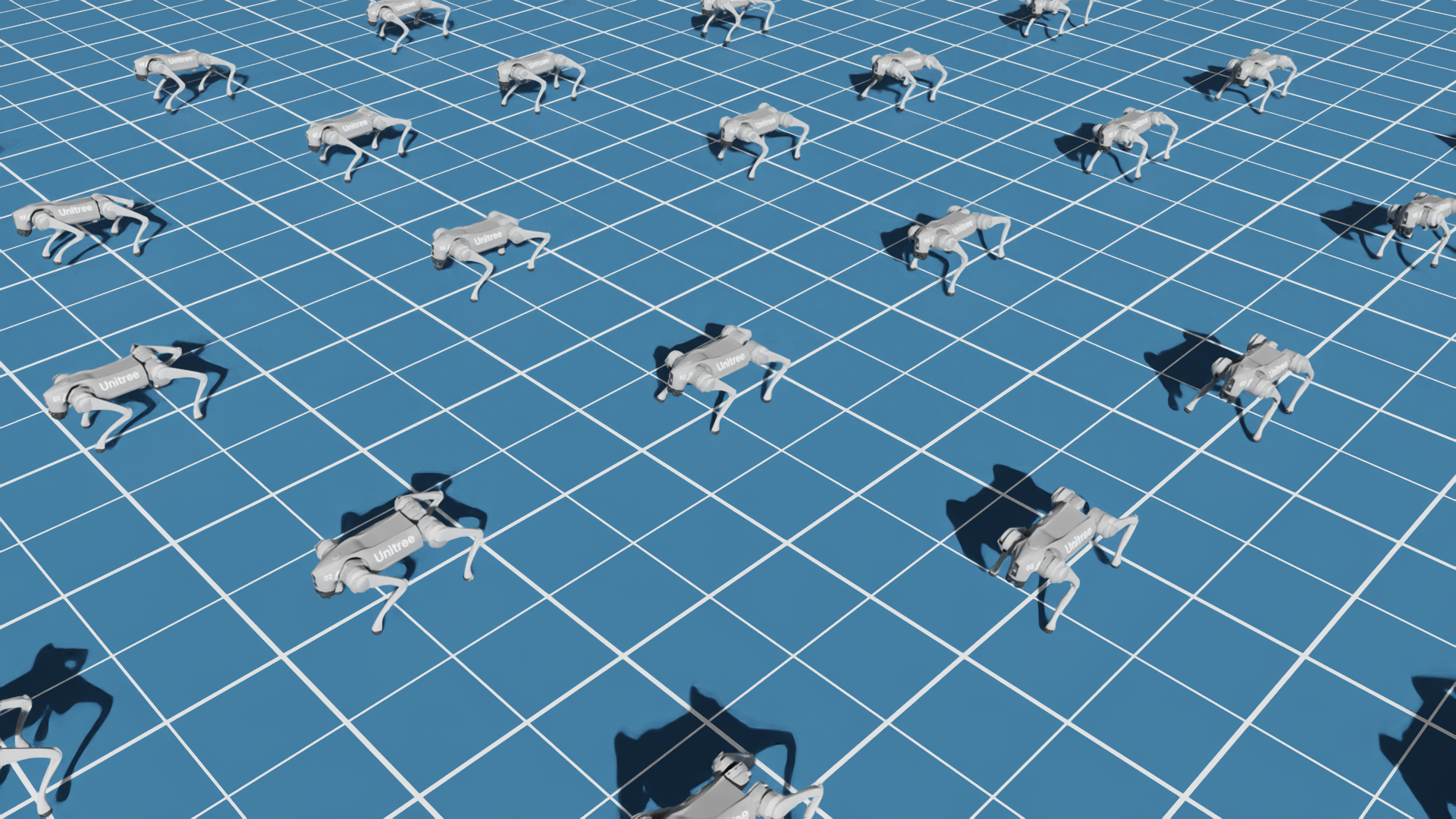}
  \caption{Parallel multi-environment simulation of the legged locomotion test.}
  \label{fig:isaacipc_qual_legged}
\end{figure}

\begin{figure}[t]
  \centering
  \includegraphics[width=\linewidth]{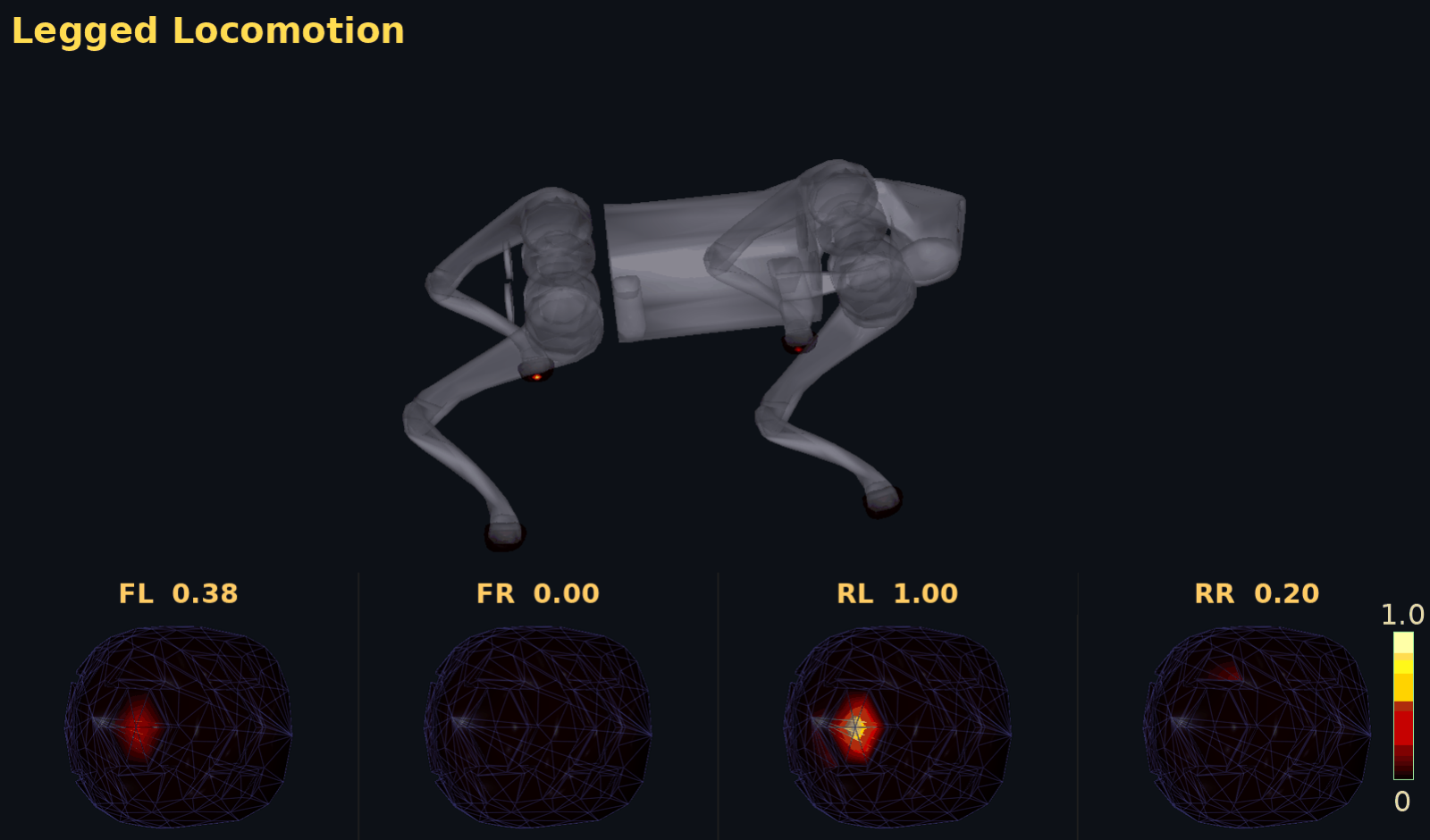}
  \caption{Contact force distribution on the four foot pads of the legged robot.}
  \label{fig:contact_distribution_foot}
\end{figure}

To ensure that the qualitative behavior of the system is governed by transparent and interpretable kinematic commands, we use a manually designed open-loop trot controller as the motion input. The controller generates an alternating diagonal-leg trot gait and plans D-shaped foot trajectories in the sagittal plane. Fig.~\ref{fig:contact_distribution_foot} visualizes the distribution of contact forces between the feet and the ground. The same setup also supports multi-environment parallel execution with domain randomization over foot elastic modulus, friction coefficient, and controller parameters, enabled by the subscene-based IsaacIPC architecture (see Fig.~\ref{fig:isaacipc_qual_legged}). 
Single-environment reset further enables selective reset, providing a practical option for rigid-deformable coupled reinforcement learning environments.

\emph{Dexterous in-hand manipulation.}
The Sharpa Wave dexterous hand~\cite{SharpaRobotics2026}, shown in the middle panel of Fig.~\ref{fig:teaser}, is demonstrated on an in-hand ball-rotation task. As in the legged locomotion example, the rigid links and joint constraints are handled by libuipc, while the five green elastomer fingertips and the manipulated ball are modeled with FEM. The target trajectories for the hand's 22 actuated degrees of freedom are obtained from rollouts of a trained IsaacLab in-hand manipulation policy.

\begin{figure}[t]
  \centering
  \includegraphics[width=\linewidth]{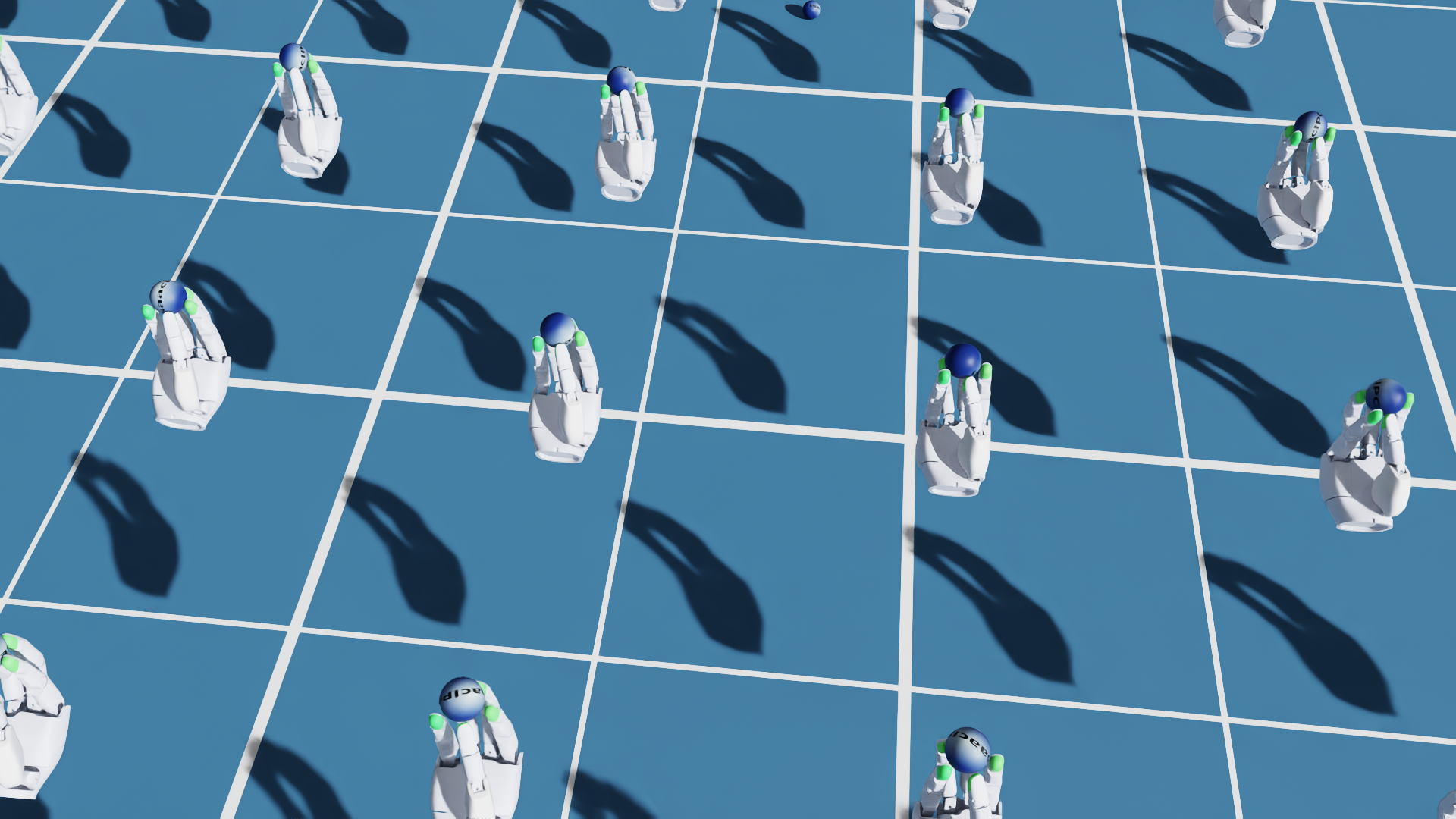}
  \caption{Parallel multi-environment simulation of dexterous hand manipulation test.}
  \label{fig:isaacipc_qual_dexhand}
\end{figure}

\begin{figure}[t]
  \centering
  \includegraphics[width=\linewidth]{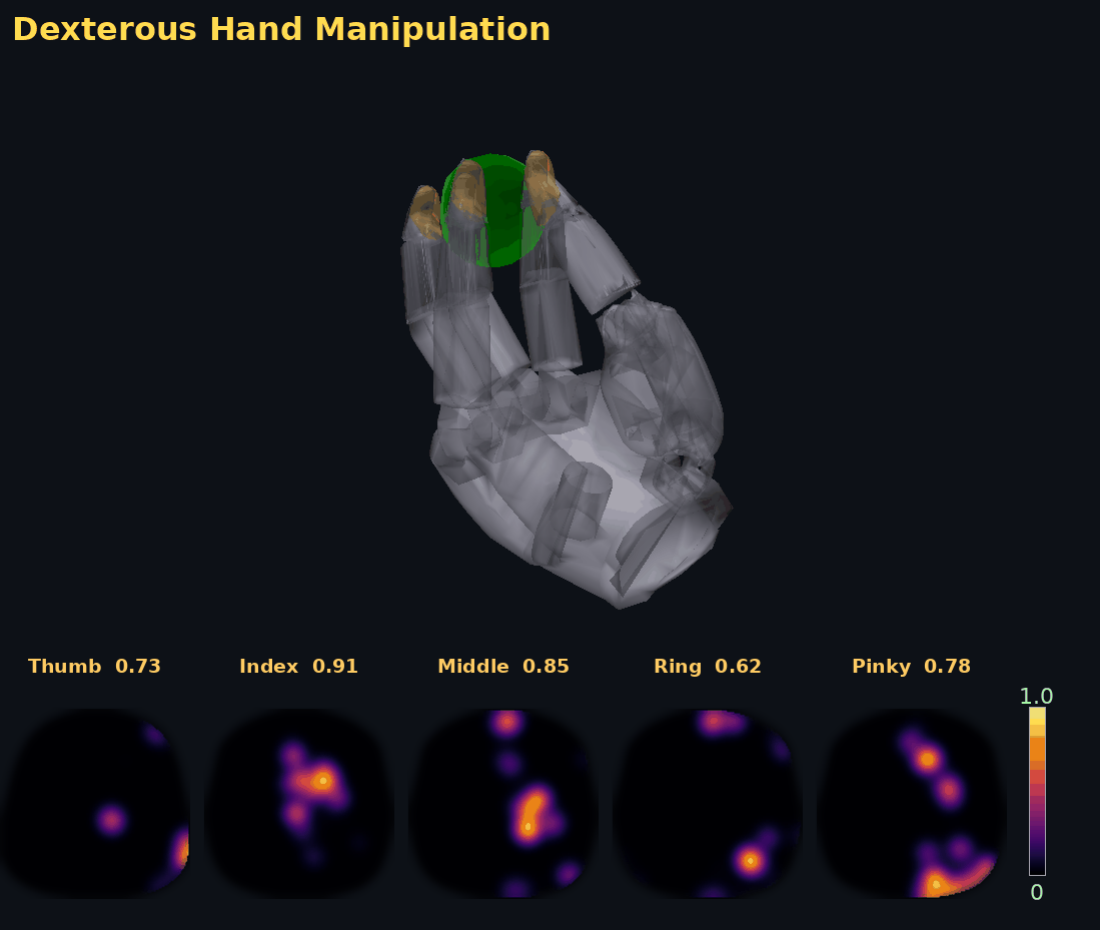}
  \caption{Contact force distribution on the five fingertips of the dexterous hand.}
  \label{fig:contact_distribution_hand}
\end{figure}

Fig.~\ref{fig:contact_distribution_hand} visualizes the fingertip contact force distribution, and Fig.~\ref{fig:isaacipc_qual_dexhand} shows the corresponding domain randomized multi environment simulation. This example illustrates how IsaacIPC can support high-quality data collection for dexterous in-hand manipulation.

\begin{figure}[t]
  \centering
  \includegraphics[width=\linewidth]{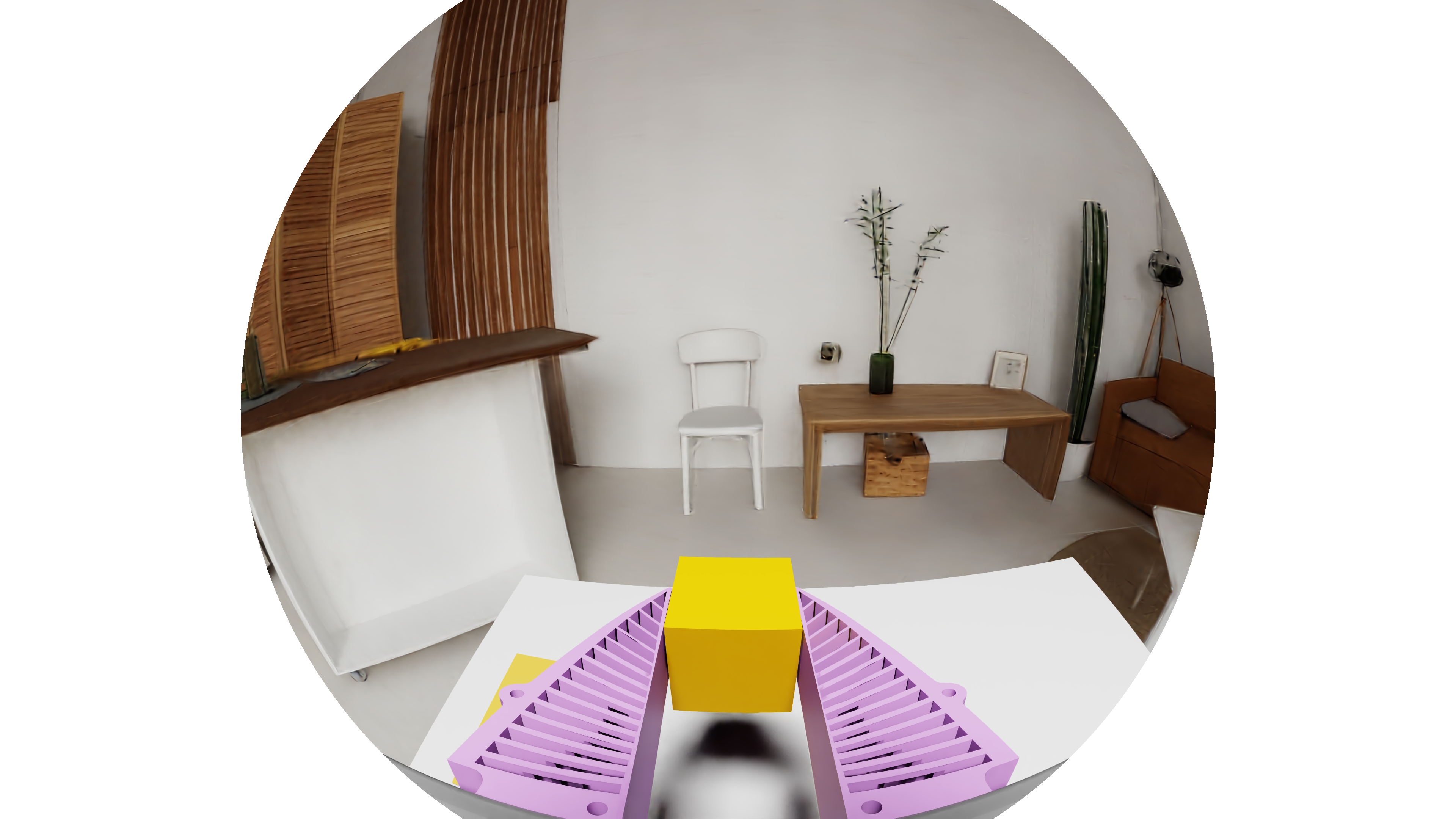}
  \caption{Fisheye view of UMI pick-and-place manipulation with a rigid cube.}
  \label{fig:isaacipc_qual_umi}
\end{figure}

\begin{figure}[t]
  \centering
  \includegraphics[width=\linewidth]{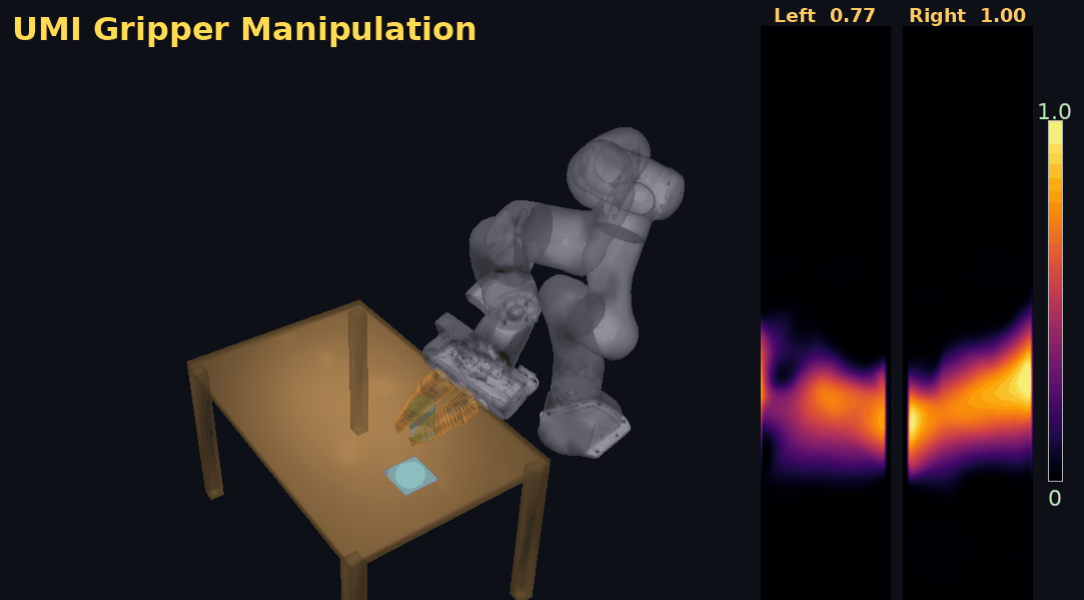}
  \caption{Contact force distribution on the inner flat surfaces of the UMI gripper.}
  \label{fig:contact_distribution_cylinder}
\end{figure}

\emph{UMI gripper manipulation.}
The Universal Manipulation Interface (UMI)~\cite{chi2024universal} facilitates large-scale manipulation data collection for robot policy training. 
We model the UMI gripper in IsaacIPC. In this test, the Franka robotic arm inverse kinematics is solved by IsaacSim, and the UMI gripper is driven one-way by the arm. 
Fig.~\ref{fig:isaacipc_qual_umi} shows a fisheye view of rigid-cube pick-and-place manipulation, with the scene background sourced from LeHome~\cite{li2026lehomesimulationenvironmentdeformable}. 
To expose richer contact details on the inner gripper surfaces, where piezoresistive tactile sensors are commonly mounted, we additionally show a softer-cylinder grasping example in Fig.~\ref{fig:contact_distribution_cylinder}. 
This setup can also be used to evaluate robot learning policies.

\section{Conclusion}\label{sec:conclusion}

We presented IsaacIPC, a robotic simulation framework that connects libuipc with IsaacSim/Lab. Through a dual-mesh mapper, IsaacIPC
computes contact and deformation on simulation meshes while preserving the
visual meshes, material bindings, USD\,/\,Fabric representation, and parallel
environment infrastructure used by Omniverse based rendering and robot
simulation. This design is not tied to a single contact model and can be
extended to other physics backends.

We also introduced geometric mortar contact potential (GMCP), a mortar-based contact formulation for tactile
simulation. GMCP constructs a barrier potential from contact samples on the
tactile surface, providing a direct discretization of contact pressure on the
sensor side. The contact patch and Hertzian contact tests validate its behavior
on normal contact-pressure transfer, and the robotic examples demonstrate how
IsaacIPC can be used in rigid--deformable contact-rich systems.

Several limitations remain. High-fidelity robotic simulation always have to balance
accuracy, robustness, and efficiency, especially when scaling rigid--deformable
contact simulation to real-time or near-real-time reinforcement-learning
settings. The current GMCP formulation mainly focuses on normal contact
pressure; tangential traction, friction, stick--slip transitions, and shear
deformation are important for tactile sensing and should be considered. 
Mortar formulations also depend on reliable projection, clipping,
and quadrature on the contact interface, and can become more involved near
sharp features, rapidly changing contact topology, or highly nonmatching
meshes. Exploring alternative pressure-transfer schemes, adaptive quadrature,
or higher-order contact representations may further improve accuracy, robustness, and efficiency.

Future work will also broaden IsaacIPC beyond the current rigid--deformable
examples. Promising directions include coupling GPU-accelerated IPC with
additional physical effects such as fluids, fracture, and other multiphysics
phenomena; integrating asset generation and scene reconstruction pipelines; and
using large-scale simulation to generate tactile and force-feedback data for
pretraining and post-training embodied foundation models.


\bibliographystyle{ACM-Reference-Format}
\bibliography{sample-base}

@String{Computing = "Computing" }

@String{Computer = "{IEEE} Computer" }

@String{Springer = "Springer-Verlag" }

@ArtifactSoftware{R,
    title = {R: A Language and Environment for Statistical Computing},
    author = {{R Core Team}},
    organization = {R Foundation for Statistical Computing},
    address = {Vienna, Austria},
    year = {2019},
    url = {https://www.R-project.org/},
}

@book{johnson1987contact,
  title={Contact mechanics},
  author={Johnson, Kenneth Langstreth},
  year={1987},
  publisher={Cambridge university press}
}

@book{laursen2003computational,
  title={Computational contact and impact mechanics: fundamentals of modeling interfacial phenomena in nonlinear finite element analysis},
  author={Laursen, Tod A},
  year={2003},
  publisher={Springer Science \& Business Media}
}

@article{Wohlmuth2000DualMortar,
  author  = {Wohlmuth, Barbara I.},
  title   = {A Mortar Finite Element Method Using Dual Spaces for the Lagrange Multiplier},
  journal = {SIAM Journal on Numerical Analysis},
  volume  = {38},
  number  = {3},
  pages   = {989--1012},
  year    = {2000},
  doi     = {10.1137/S0036142999350929},
}

@article{Puso2004Mortar,
  author  = {Puso, Michael A. and Laursen, Tod A.},
  title   = {A Mortar Segment-to-Segment Contact Method for Large Deformation Solid Mechanics},
  journal = {Computer Methods in Applied Mechanics and Engineering},
  volume  = {193},
  number  = {6--8},
  pages   = {601--629},
  year    = {2004},
  doi     = {10.1016/j.cma.2003.10.010},
}

@article{Li2020IPC,
  author    = {Minchen Li and Zachary Ferguson and Teseo Schneider and Timothy Langlois and Denis Zorin and Daniele Panozzo and Chenfanfu Jiang and Danny M. Kaufman},
  title     = {Incremental Potential Contact: Intersection- and Inversion-free, Large-Deformation Dynamics},
  journal   = {ACM Transactions on Graphics},
  volume    = {39},
  number    = {4},
  articleno = {49},
  year      = {2020},
  month     = jul,
  numpages  = {20},
  doi       = {10.1145/3386569.3392425},
}

@article{Ferguson2021RigidIPC,
  author    = {Zachary Ferguson and Minchen Li and Teseo Schneider and Francisca Gil-Ureta and Timothy Langlois and Chenfanfu Jiang and Denis Zorin and Danny M. Kaufman and Daniele Panozzo},
  title     = {Intersection-free Rigid Body Dynamics},
  journal   = {ACM Transactions on Graphics},
  volume    = {40},
  number    = {4},
  articleno = {183},
  year      = {2021},
  month     = aug,
  numpages  = {16},
  doi       = {10.1145/3450626.3459802},
}

@inproceedings{Kim2022IPCGraspSim,
  author    = {Chung Min Kim and Michael Danielczuk and Isabella Huang and Ken Goldberg},
  title     = {{IPC-GraspSim}: Reducing the Sim2Real Gap for Parallel-Jaw Grasping with the Incremental Potential Contact Model},
  booktitle = {Proceedings of the IEEE International Conference on Robotics and Automation (ICRA)},
  year      = {2022},
  pages     = {6180--6187},
}

@article{Lan2022PDGPU,
  author    = {Lei Lan and Guanqun Ma and Yin Yang and Changxi Zheng and Minchen Li and Chenfanfu Jiang},
  title     = {Penetration-free Projective Dynamics on the {GPU}},
  journal   = {ACM Transactions on Graphics},
  volume    = {41},
  number    = {4},
  articleno = {69},
  year      = {2022},
  month     = jul,
  numpages  = {16},
  doi       = {10.1145/3528223.3530069},
}

@article{Lan2022ABD,
  author    = {Lei Lan and Danny M. Kaufman and Minchen Li and Chenfanfu Jiang and Yin Yang},
  title     = {Affine Body Dynamics: Fast, Stable and Intersection-free Simulation of Stiff Materials},
  journal   = {ACM Transactions on Graphics},
  volume    = {41},
  number    = {4},
  articleno = {67},
  year      = {2022},
  month     = jul,
  numpages  = {14},
  doi       = {10.1145/3528223.3530064},
}

@article{Chen2022UnifiedIPC,
  author    = {Yunuo Chen and Minchen Li and Lei Lan and Hao Su and Yin Yang and Chenfanfu Jiang},
  title     = {A Unified Newton Barrier Method for Multibody Dynamics},
  journal   = {ACM Transactions on Graphics},
  volume    = {41},
  number    = {4},
  articleno = {66},
  year      = {2022},
  month     = jul,
  numpages  = {14},
  doi       = {10.1145/3528223.3530076},
}

@inproceedings{Ferguson2023HighOrderIPC,
  author    = {Zachary Ferguson and Pranav Jain and Denis Zorin and Teseo Schneider and Daniele Panozzo},
  title     = {High-Order Incremental Potential Contact for Elastodynamic Simulation on Curved Meshes},
  booktitle = {Special Interest Group on Computer Graphics and Interactive Techniques Conference Proceedings (SIGGRAPH '23)},
  year      = {2023},
  month     = aug,
  address   = {Los Angeles, CA, USA},
  doi       = {10.1145/3588432.3591488},
}

@article{Ferguson2023ITR,
  author    = {Zachary Ferguson and Teseo Schneider and Danny M. Kaufman and Daniele Panozzo},
  title     = {In-Timestep Remeshing for Contacting Elastodynamics},
  journal   = {ACM Transactions on Graphics},
  volume    = {42},
  number    = {4},
  articleno = {145},
  year      = {2023},
  month     = aug,
  numpages  = {15},
  doi       = {10.1145/3592428},
}

@misc{Li2023ConvergentIPC,
  author        = {Minchen Li and Zachary Ferguson and Teseo Schneider and Timothy Langlois and Denis Zorin and Daniele Panozzo and Chenfanfu Jiang and Danny M. Kaufman},
  title         = {Convergent Incremental Potential Contact},
  year          = {2023},
  eprint        = {2307.15908},
  archivePrefix = {arXiv},
  primaryClass  = {math.NA},
}

@article{Du2024ZeMa,
  author    = {Wenxin Du and Siqiong Yao and Xinlei Wang and Yuhang Xu and Wenqiang Xu and Cewu Lu},
  title     = {Intersection-free Robot Manipulation with Soft-Rigid Coupled Incremental Potential Contact},
  journal   = {IEEE Robotics and Automation Letters},
  year      = {2024},
}

@article{Du2024TacIPC,
  author    = {Wenxin Du and Wenqiang Xu and Jieji Ren and Zhenjun Yu and Cewu Lu},
  title     = {{TacIPC}: Intersection- and Inversion-free {FEM}-based Elastomer Simulation for Optical Tactile Sensors},
  journal   = {IEEE Robotics and Automation Letters},
  year      = {2024},
}

@article{Huang2024GIPC,
  author    = {Kemeng Huang and Floyd M. Chitalu and Huancheng Lin and Taku Komura},
  title     = {{GIPC}: Fast and Stable {Gauss-Newton} Optimization of {IPC} Barrier Energy},
  journal   = {ACM Transactions on Graphics},
  volume    = {43},
  number    = {2},
  year      = {2024},
  month     = jan,
  numpages  = {18},
  doi       = {10.1145/3643028},
}

@article{Guo2024BarrierAL,
  author    = {Dewen Guo and Minchen Li and Yin Yang and Sheng Li and Guoping Wang},
  title     = {Barrier-Augmented Lagrangian for {GPU}-based Elastodynamic Contact},
  journal   = {ACM Transactions on Graphics},
  volume    = {43},
  number    = {6},
  articleno = {225},
  year      = {2024},
  month     = nov,
  numpages  = {17},
  doi       = {10.1145/3687988},
}

@inproceedings{Shen2024PNCGIPC,
  author    = {Xing Shen and Runyuan Cai and Mengxiao Bi and Tangjie Lv},
  title     = {Preconditioned Nonlinear Conjugate Gradient Method for Real-time Interior-point Hyperelasticity},
  booktitle = {Special Interest Group on Computer Graphics and Interactive Techniques Conference Papers (SIGGRAPH '24)},
  year      = {2024},
  month     = jul,
  address   = {Denver, CO, USA},
  numpages  = {11},
  doi       = {10.1145/3641519.3657490},
}

@misc{Du2024EmbeddedIPC,
  author        = {Wenxin Du and Chang Yu and Siyu Ma and Ying Jiang and Zeshun Zong and Yin Yang and Joe Masterjohn and Alejandro Castro and Xuchen Han and Chenfanfu Jiang},
  title         = {Embedded {IPC}: Fast and Intersection-free Simulation in Reduced Subspace for Robot Manipulation},
  year          = {2024},
  eprint        = {2409.16385},
  archivePrefix = {arXiv},
  primaryClass  = {cs.RO},
}

@inproceedings{Nguyen2024TacEx,
  author    = {Duc Huy Nguyen and Tim Schneider and Guillaume Duret and Alap Kshirsagar and Boris Belousov and Jan Peters},
  title     = {{TacEx}: {GelSight} Tactile Simulation in {Isaac Sim} -- Combining Soft-Body and Visuotactile Simulators},
  booktitle = {8th Conference on Robot Learning (CoRL 2024)},
  year      = {2024},
  address   = {Munich, Germany},
}

@article{Huang2025StiffGIPC,
  author    = {Kemeng Huang and Xinyu Lu and Huancheng Lin and Taku Komura and Minchen Li},
  title     = {{StiffGIPC}: Advancing {GPU} {IPC} for Stiff Affine-Deformable Simulation},
  journal   = {ACM Transactions on Graphics},
  year      = {2025},
}

@article{Huang2025GCP,
  author    = {Zizhou Huang and Maxwell Paik and Zachary Ferguson and Daniele Panozzo and Denis Zorin},
  title     = {Geometric Contact Potential},
  journal   = {ACM Transactions on Graphics},
  volume    = {44},
  number    = {4},
  year      = {2025},
  month     = aug,
  numpages  = {24},
  doi       = {10.1145/3731142},
}

@misc{Li2025Taccel,
  author        = {Yuyang Li and Wenxin Du and Chang Yu and Puhao Li and Zihang Zhao and Tengyu Liu and Chenfanfu Jiang and Yixin Zhu and Siyuan Huang},
  title         = {Taccel: Scaling Up Vision-based Tactile Robotics via High-performance {GPU} Simulation},
  year          = {2025},
  eprint        = {2504.12908},
  archivePrefix = {arXiv},
  primaryClass  = {cs.RO},
}

@article{Zheng2026ALIPC,
  author    = {Juntian Zheng and Zhaofeng Luo and Minchen Li},
  title     = {Robust and Efficient Penetration-Free Elastodynamics without Barriers},
  journal   = {ACM Transactions on Graphics},
  year      = {2026},
  doi       = {10.1145/3811035},
}

@misc{Chen2026UniVTAC,
  author        = {Baijun Chen and Weijie Wan and Tianxing Chen and Xianda Guo and Congsheng Xu and Yuanyang Qi and Haojie Zhang and Longyan Wu and Tianling Xu and Zixuan Li and Yizhe Wu and Rui Li and Xiaokang Yang and Ping Luo and Wei Sui and Yao Mu},
  title         = {{UniVTAC}: A Unified Simulation Platform for Visuo-Tactile Manipulation Data Generation, Learning, and Benchmarking},
  year          = {2026},
  eprint        = {2602.10093},
  archivePrefix = {arXiv},
  primaryClass  = {cs.RO},
}

@misc{Yan2026Tac2Real,
  author        = {Ningyu Yan and Shuai Wang and Xing Shen and Hui Wang and Hanqing Wang and Yang Xiang and Jiangmiao Pang},
  title         = {{Tac2Real}: Reliable and {GPU} Visuotactile Simulation for Online Reinforcement Learning and Zero-Shot Real-World Deployment},
  year          = {2026},
  eprint        = {2603.28475},
  archivePrefix = {arXiv},
  primaryClass  = {cs.RO},
}

@misc{Zhang2026MPNCGIPC,
  author        = {Yu Zhang and Xing Shen and Kemeng Huang and Wei Chen and Yin Yang and Taku Komura and Tiantian Liu and Xingang Pan},
  title         = {An Efficient Multilevel Preconditioned Nonlinear Conjugate Gradient Method for Incremental Potential Contact},
  year          = {2026},
  eprint        = {2604.19892},
  archivePrefix = {arXiv},
  primaryClass  = {cs.GR},
}

@article{Sutherland1974,
  author    = {Sutherland, Ivan E. and Hodgman, Gary W.},
  title     = {Reentrant polygon clipping},
  year      = {1974},
  journal   = {Commun. ACM},
  volume    = {17},
  number    = {1},
  pages     = {32--42},
  month     = jan,
  publisher = {Association for Computing Machinery},
  doi       = {10.1145/360767.360802},
  issn      = {0001-0782},
}

@article{farah2018mortar,
  author    = {Farah, P. and Wall, W. A. and Popp, A.},
  title     = {A mortar finite element approach for point, line, and surface contact},
  journal   = {International Journal for Numerical Methods in Engineering},
  year      = {2018},
  volume    = {114},
  number    = {3},
  pages     = {255--291},
  doi       = {10.1002/nme.5743},
}

@article{Si2015TetGen,
  author    = {Si, Hang},
  title     = {{TetGen}, a {Delaunay}-Based Quality Tetrahedral Mesh Generator},
  journal   = {ACM Transactions on Mathematical Software},
  year      = {2015},
  volume    = {41},
  number    = {2},
  pages     = {11:1--11:36},
  doi       = {10.1145/2629697},
}

@article{Capell2002Skeleton,
  author    = {Capell, Steve and Green, Seth and Curless, Brian and Duchamp, Tom and Popovi\'{c}, Zoran},
  title     = {Interactive Skeleton-Driven Dynamic Deformations},
  journal   = {ACM Transactions on Graphics},
  year      = {2002},
  volume    = {21},
  number    = {3},
  pages     = {586--593},
  doi       = {10.1145/566654.566622},
}

@misc{NVIDIA_PhysX,
  author       = {{NVIDIA Corporation}},
  title        = {{PhysX SDK}},
  howpublished = {\url{https://developer.nvidia.com/physx-sdk}},
  year         = {2025},
}

@misc{NVIDIAWarp,
  author       = {Macklin, Miles},
  title        = {{Warp}: A High-performance Python Framework for {GPU} Simulation and Graphics},
  year         = {2022},
  month        = mar,
  note         = {{NVIDIA GPU Technology Conference (GTC)}},
  howpublished = {\url{https://github.com/NVIDIA/warp}},
}

@misc{NVIDIAOmniverse2026,
  author       = {{NVIDIA}},
  title        = {{NVIDIA Omniverse}},
  howpublished = {\url{https://www.nvidia.com/en-us/omniverse/}},
  year         = {2026},
  note         = {Accessed: 2026-05-22},
}

@misc{lai2025hunyuan3d25,
  title        = {Hunyuan3D 2.5: Towards High-Fidelity 3D Assets Generation with Ultimate Details},
  author       = {{Tencent Hunyuan3D Team}},
  year         = {2025},
  eprint       = {2506.16504},
  archivePrefix= {arXiv},
  primaryClass = {cs.CV},
  url          = {https://arxiv.org/abs/2506.16504},
}

@misc{Genesis,
  author       = {{Genesis Authors}},
  title        = {Genesis: A Generative and Universal Physics Engine for Robotics and Beyond},
  month        = {December},
  year         = {2024},
  howpublished = {\url{https://github.com/Genesis-Embodied-AI/Genesis}},
}

@article{mittal2025isaac,
  title   = {Isaac lab: A gpu-accelerated simulation framework for multi-modal robot learning},
  author  = {Mittal, Mayank and Roth, Pascal and Tigue, James and Richard, Antoine and Zhang, Octi and Du, Peter and Serrano-Munoz, Antonio and Yao, Xinjie and Zurbr{\"u}gg, Ren{\'e} and Rudin, Nikita and others},
  journal = {arXiv preprint arXiv:2511.04831},
  year    = {2025},
}

@article{chi2024universal,
  title   = {Universal Manipulation Interface: In-the-Wild Robot Teaching Without In-the-Wild Robots},
  author  = {Chi, Cheng and Xu, Zhenjia and Pan, Chuer and Cousineau, Eric and Burchfiel, Benjamin and Feng, Siyuan and Tedrake, Russ and Song, Shuran},
  journal = {arXiv preprint arXiv:2402.10329},
  year    = {2024},
}

@inproceedings{yuan2015measurement,
  title     = {Measurement of Shear and Slip with a {GelSight} Tactile Sensor},
  author    = {Yuan, Wenzhen and Li, Rui and Srinivasan, Mandayam A. and Adelson, Edward H.},
  booktitle = {2015 IEEE International Conference on Robotics and Automation (ICRA)},
  pages     = {304--311},
  year      = {2015},
  organization = {IEEE},
}

@misc{li2026lehomesimulationenvironmentdeformable,
  title         = {LeHome: A Simulation Environment for Deformable Object Manipulation in Household Scenarios},
  author        = {Zeyi Li and Yushi Yang and Shawn Xie and Kyle Xu and Tianxing Chen and Yuran Wang and Zhenhao Shen and Yan Shen and Yue Chen and Wenjun Li and Yukun Zheng and Chaorui Zhang and Siyi Lin and Fei Teng and Hongjun Yang and Ming Chen and Steve Xie and Ruihai Wu},
  year          = {2026},
  eprint        = {2604.22363},
  archivePrefix = {arXiv},
  primaryClass  = {cs.RO},
  url           = {https://arxiv.org/abs/2604.22363},
}

@misc{SharpaRobotics2026,
  author       = {{Sharpa Robotics}},
  title        = {{Sharpa Robotics GitHub Organization}},
  howpublished = {\url{https://github.com/sharpa-robotics}},
  year         = {2026},
  note         = {Accessed: 2026-05-20},
}

@misc{UnitreeRobotics2026,
  author       = {{Unitree Robotics}},
  title        = {{Unitree Robotics GitHub Organization}},
  howpublished = {\url{https://github.com/unitreerobotics}},
  year         = {2026},
  note         = {Accessed: 2026-05-20},
}

@misc{libuipc,
  author       = {{spiriMirror}},
  title        = {{libuipc}},
  howpublished = {\url{https://github.com/spiriMirror/libuipc}},
  year         = {2026},
  note         = {Accessed: 2026-05-21},
}

@book{wriggers2006computational,
  title     = {Computational Contact Mechanics},
  author    = {Wriggers, Peter and Laursen, Tod A},
  volume    = {2},
  year      = {2006},
  publisher = {Springer}
}
\end{document}